\documentclass[10pt,twocolumn,letterpaper]{article}

\usepackage{wacv}
\usepackage{times}
\usepackage{epsfig}
\usepackage{graphicx}
\usepackage{amsmath}
\usepackage{amssymb}

\usepackage{dsfont}

\DeclareMathOperator*{\argmin}{arg\,min}

\wacvfinalcopy 

\ifwacvfinal\fi
\begin{document}

\title{Vehicle Re-Identification: an Efficient Baseline Using Triplet Embedding}

\author{Ratnesh Kumar \hspace{2cm} Edwin Weill \hspace{2cm} Farzin Aghdasi \hspace{2cm} Parthasarathy Sriram \\
NVIDIA\\
{\tt\small \{ ratneshk,eweill \} @nvidia.com}
}

\maketitle

\begin{abstract}\footnote{Accepted at IEEE IJCNN 2019. This arxiv version also adds experiment on recently proposed datasets.}
  In this paper we tackle the problem of vehicle re-identification in a camera
  network utilizing triplet embeddings. Re-identification is the problem of matching appearances of objects across
  different cameras.
  With the proliferation of surveillance cameras enabling smart and safer cities, there is an ever-increasing
  need to re-identify vehicles across cameras. Typical challenges arising in smart city scenarios include variations of viewpoints,
  illumination and self occlusions.
  Most successful approaches for re-identification
  involve (deep) learning an embedding space such that the vehicles of same identities are projected closer to one another, compared to the vehicles representing
  different identities. Popular loss functions for learning an embedding (space) include \emph{contrastive} or
  \emph{triplet} loss. In this paper we provide an extensive evaluation of these losses applied to vehicle
  re-identification and demonstrate that using the best
  practices for learning embeddings outperform most of the previous approaches proposed in the vehicle re-identification literature. Compared to most existing state-of-the-art approaches, our approach is simpler and more straightforward for training utilizing only identity-level annotations, along with one of the smallest published embedding dimensions for efficient inference. Furthermore in this work we introduce a formal evaluation of a triplet sampling variant (\emph{batch sample}) into the re-identification literature.
\end{abstract}

\section{Introduction}
Matching appearances of objects across multiple cameras is an important problem for many computer vision applications, \emph{e.g.} object retrieval and object identification.
This problem of object re-identification is closely related to object recognition and fine grained
classification. In the realm of video understanding, most higher level algorithms such as event recognition and anomaly detection rely upon \emph{Multiple Camera Multiple Object Tracking} (MC-MOT). An important
component for a MC-MOT is an \emph{object verification} (\emph{i.e.} re-identification) module for expressing \emph{confidence} to associate objects across
multiple videos \cite{Ristani2018}. Re-identification approaches can also be used in a single camera setup, wherein the task would be to
determine if the same object has re-appeared in the scene \cite{Kumar2014a, Wojke2018, Tang2017}.

\begin{figure}[t]
\centering
\includegraphics[width=0.115\textwidth]{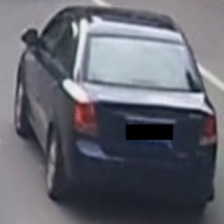}
\includegraphics[width=0.115\textwidth]{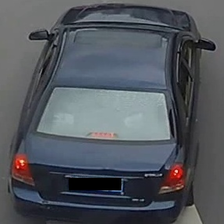}
\includegraphics[width=0.115\textwidth]{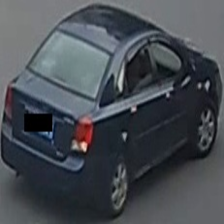}
\includegraphics[width=0.115\textwidth]{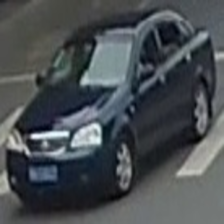}
\\
\includegraphics[width=0.115\textwidth]{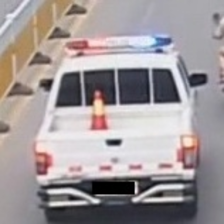}
\includegraphics[width=0.115\textwidth]{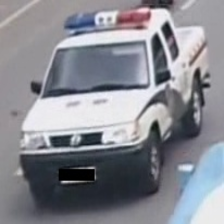}
\includegraphics[width=0.115\textwidth]{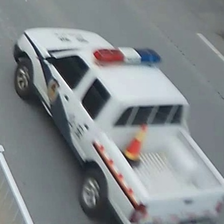}
\includegraphics[width=0.115\textwidth]{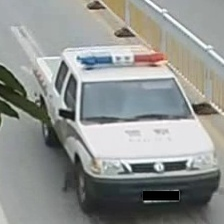}
\\
\includegraphics[width=0.115\textwidth]{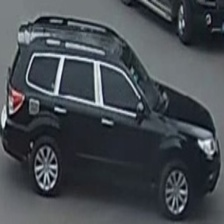}
\includegraphics[width=0.115\textwidth]{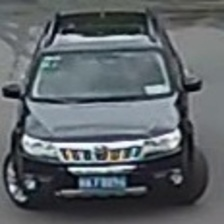}
\includegraphics[width=0.115\textwidth]{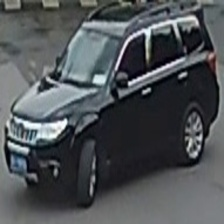}
\includegraphics[width=0.115\textwidth]{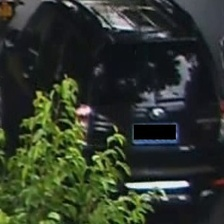}
\caption{Each row is a separate identity (samples taken from \textbf{VeRi} dataset \cite{Liu2016a}). Despite large intra-class
  variations for views, vehicle-model could be discerned from most views.}
\label{fig:intro}
\end{figure}

The task of vehicle re-identification is to identify the same vehicle across a camera network. With the deployment of camera sensors
for traffic management and smart cities, there is an imminent need to perform vehicle search from video
databases \cite{Naphade2018}. Previous works \cite{Spanhel2017,Jain2016} have shown that automatic recognition of license plates as a global unique identifier have given state-of-the-art identification performance. However in general traffic
scenes at streets, license plates are practically invisible in many views to recognize due to their top view
installations. Therefore, a vision-based re-identification has a great practical value in real world scenarios.
Re-identification of objects is challenging due to significant appearance \& viewpoint shifts, lighting variations and varied
object poses. Figure \ref{fig:intro} shows some typical challenging intra-class variations.

Compared to person and face re-identification, vehicle re-identification is a relatively under-studied
problem. A few of the unique characteristics pertaining to the problem of vehicle re-identification which make it a difficult task are:
\begin{itemize}
  \item Multiple views of the same vehicle are visually diverse and semantically (i.e. color and model) correlated, meaning that the same identity must be deduced no matter which viewpoint of the vehicle is given. 
  
  \item In real world scenarios, a re-identification system is expected to extract subtle physical cues such as the presence of dust, written marks, or dents on vehicle surfaces, to be able to distinguish between vehicles which are the same color and model. 
  \item The vehicle labels are less fine-grained than person (or face)-identity labels. Given that there are a finite
      number of vehicle colors and models, the diversity in a given dataset is less than that of a person or face
      re-identification dataset.
\end{itemize}

In order
to match appearances of objects, firstly we need to obtain an embedding for the objects, also denoted as a feature vector
or signature. A match is then performed by using a suitable distance metric expressing the closeness of two
objects in an embedding space. A good embedding should be invariant to illumination, scale and viewpoint changes. Prior to the advancements in
deep learning, most embedding learning approaches focus on handcrafting using mixture of multiple feature extractors
and/or learning suitable
ranking functions to minimize distance across objects of similar identities. Some of the notable approaches are
\cite{Tuzel2006, Bak2017, Ma2012a, Ma2013, Farenzena2010, Liao2015, Zapletal2016}.

\emph{In this paper} we focus on the embedding part of the re-identification process and make the following \textbf{contributions}:
\begin{itemize}
    \item Utilizing the recent advances in \emph{sampling} informative data points for learning embedding for the person re-identification task\cite{Hermans2017}, we extensively evaluate their application to the vehicle re-identification problem, and demonstrate state-of-the-art performance across diverse datasets on various performance metrics.
    \item We introduce a formal evaluation of a triplet sampling variant, \emph{batch sample}, into the re-identification literature.
\end{itemize}

The rest of the paper is organized as: in the following section we provide an overview of related works and the subsequent section will elaborate on triplet and contrastive
losses, including popular sampling techniques to optimize these losses. Section \ref{sec:experiments} details
on datasets and hyperparameters used for various experimental settings. Results and discussions are presented in section
\ref{sec:results}.

\section{Related Works}

In recent years with the evolution of end-to-end learning using \emph{Convolutional Neural Networks} (CNN),
significant improvements have been made in feature representations
using large amounts of training data. These approaches outperform all
previous baselines using handcrafted features. A CNN learns hierarchical image features by stacking convolutional layers
with downsampling layers. The outputs from one convolutional layer is fed to a non-linearity layer before being fed to
the subsequent convolutional layer.

\cite{BROMLEY1993} proposed one of the first approaches to learn visual
relationships using CNN. \emph{Siamese} CNN \cite{BROMLEY1993} computes an embedding space such that similar examples
have similar embeddings and vice versa. \cite{Chopra2005} uses \emph{contrastive} loss on Siamese CNN to learn embedding for
face verification. One of the recent prominent works using CNNs for learning face embedding \cite{Schroff2015} uses
\emph{triplet loss} to train a CNN for learning face embeddings for identification. While triplet loss considers three samples \emph{jointly} for computing a loss measure, contrastive
loss requires only two samples. Contrastive loss is computationally more efficient than triplet, however, several approaches
\cite{Manmatha2017,Guo2018,Ristani2018,Bai2018,Hermans2017,Hoffer2015} have
reported state-of-the-art performances using triplet loss. This superiority of triplet loss is attributed to the
additional context using three samples. Section \ref{sec:loss} in this paper elaborates on these losses.

Another method for obtaining an embedding for an object is utilizing a traditional softmax layer
\cite{Xiao2016,Kanaci2017}, wherein a fully-connected (embedding) layer is added prior to the softmax-loss layer. Each identity is considered as a
separate category and the number of categories is equal to the number of identities in the training set.  Once the network is trained using
classification loss (\emph{e.g.} cross-entropy), the classification layer is
stripped off and an embedding is obtained form the new final layer of the network. \cite{Kanaci2017} proposed a similar approach to learning vehicle
embedding based on training a network for vehicle-model classification task. Since the network is not directly trained on
embedding or metric learning loss, usually the performance of such a network is poor when compared to networks incorporating embedding loss. Cross entropy loss ensures separability of features but the features may not be
discriminative enough for separating unseen identities. Furthermore learning becomes computationally prohibitive when considering datasets
of \emph{e.g.} $10^6$ identities. Some recent works \cite{Wojke2018,Rippel2015,Shen2015} unify
classification loss with metric learning.

\noindent{\textbf{Vehicle Classification}}: Fine grained vehicle classification is a closely related problem to vehicle re-identification. Notable works for
vehicle classification are \cite{Liao2015a,Gu2013,Hu2017,Lin2014,Sochor2016,Luo2015}. The general task is to predict
vehicle \emph{model}, \emph{e.g.} BMW-i3-2016, Toyota-Camry-1996. Vehicle
re-identification is a relatively finer grained problem than vehicle-model classification: a re-identification approach should be
able to extract visual differences between two vehicles belonging to the same model category. The
visual differences could include subtle cosmetic and color differences making this problem more
difficult. Furthermore a re-identification method is expected to work without any \emph{a priori} knowledge of all
possible vehicle models in
the city or a geographical entity.

\noindent{\textbf{Vehicle Re-identification}}: Some notable
approaches prior to deep learning are \cite{Liu2014,Zapletal2016}. Popular deep learning approaches for vehicle re-identification are
\cite{Wang2018,Liu2016b,Yan2017,Bai2018,Liu2016a,Guo2018,Shen2017,Wang2017,Zhou2018,Kanaci2017,Zhou2018a}. \cite{Liu2016a} proposed fusion of handcrafted
features \emph{e.g.} color, texture along with high level attribute feature obtained using CNN. \cite{Wang2018} proposed
a progressive refinement approach to searching query vehicles. A list of candidates is
obtained for a query using embeddings from a siamese-CNN trained using contrastive loss.
This list is then pruned using a siamese network to match license
plates. In order to get reliable query for visually similar vehicles, authors factor in the
usage of spatio-temporal distance comparison in addition to visual embedding distances.

\cite{Guo2018} presents a structured deep learning loss comprising a classification loss term (based on
vehicle model) as well as coarse and fine grained ranking terms. \cite{Liu2016b} proposed a
modification of triplet loss by replacing anchor samples with corresponding class center in order to suppress effects of
using poor anchors. Furthermore the deep model is trained for both vehicle model classification and identity labels in
a multi-level process. \cite{Yan2017} focuses on the relationship between different vehicle images as multiple grains by using diverse vehicle
attributes. The authors proposed ranking methods incorporated into multi-grain classification.

In a recent work \cite{Bai2018}, the authors propose to include group-based sub-clustering in a
triplet loss framework. This helps in explicitly dealing with intra-class variations of vehicle identification
problem. During training an online grouping method is used to cluster samples within each identity into disparate clusters. The
authors demonstrate state-of-the-art results in different datasets.

\cite{Zhou2018} proposes to use a view-point
synthesis approach to predict embedding for unknown views given a true view image. These synthetic embeddings for
unknown views are generated using bi-directional LSTM \cite{Hochreiter1997}. The complete network is trained using a combination of contrastive,
reconstruction and
generative adversarial loss \cite{Goodfellow2014}. Similar to the objective of \cite{Zhou2018} for inferring a global
feature vector using view-synthesis, authors in \cite{Zhou2018a} propose a \emph{viewpoint attentive multi-view}
framework. Utilizing attentive \cite{Mnih2014} and adversarial loss, authors transform a single view feature into a global multi-view feature
representation.

\cite{Wang2017} develops
a framework utilizing keypoint annotations on vehicles to learn viewpoint invariant features from a CNN. To further
enhance the retrieval of matching vehicles the authors use probabilistic spatio-temporal regularization using random
variables representing camera transition probabilities. The authors demonstrate superior results by adding this
regularization during retrieval procedure. \cite{Shen2017} formulate these \emph{camera transition} probabilities by generating
proposals of path (trajectories) and employing a LSTM and Siamese CNN to obtain a robust re-identification performance.

\section{Loss functions for embedding}\label{sec:loss}
For a reliable re-identification of objects, the following are some desired characteristics of an embedding function:
\begin{itemize}
    \setlength{\itemsep}{1pt}
  \item An embedding should be invariant to viewpoints, illumination and shape changes to the object.
  \item For a practical application deployment, computation of embedding and ranking should be
  efficient.
\end{itemize}
Consider a dataset $\mathcal{X} = \{(x_i, y_i)\}_{i=1}^{N}$ of $N$ training images $x_i \in \mathds{R}^D$ and their
corresponding class labels $y_i \in \{1 \cdots C\} $.
Re-identification approaches aim to learn an embedding $f(x; \theta) : \mathds{R}^D \rightarrow \mathds{R}^F$ to
map images in $\mathds{R}^D$ onto a feature (embedding) space in $\mathds{R}^F$ such that images of similar identity are
metrically close in this feature space. $\theta$ corresponds to the parameters of the learning function.
\begin{equation}
   \theta^* = \underset{\theta}  {\argmin} ~~{\mathcal{L}(f(\theta, \mathcal{X}))}
  \label{eq:loss}
  \end{equation}

Let $D(x_i, x_j) : \mathds{R^F}  \times  \mathds{R^F} \rightarrow \mathds{R}$ be a metric measuring distance of images
$x_i$ and $x_j$ in embedding space. For simplicity we drop the input labels and denote $D(x_i, x_j)$ as
$D_{ij}$. $y_{ij} = 1$ is both samples $i$ and $j$  belong to the same class and $y_{ij} = 0$ indicates samples of
different classes.

\subsection{Contrastive Loss}
Contrastive loss \eqref{eq:contrastive} was employed in \cite{Chopra2005} for the face verification problem, wherein the objective is to
verify if two presented faces belong to the same identity. This discriminative loss
directly optimizes \eqref{eq:loss} by encouraging all similar class distances to approach 0 while keeping all dis-similar
class distances to be above a pre-defined threshold $\alpha$.

\begin{equation}
  {l}_{contrastive}(i,j) = y_{ij}D_{ij}^2 + (1-y_{ij})[\alpha - D_{ij}^2]_+
  \label{eq:contrastive}
\end{equation}

Notice that the choice of $\alpha$ is same for all dissimilar classes. This implies that for dissimilar identities, visually diverse classes are
embedded in the same feature space as the visually similar ones. This assumption is stricter when compared to
triplet loss, and restricts the structure of the embedding manifold thereby impairing discriminative learning.
The training complexity is $O(N^2)$ for a dataset
of $N$ samples.

\subsection{Triplet Loss}
Inspired from the seminal work on metric learning for nearest neighbor classification by \cite{Weinberger2009}, \emph{facenet} \cite{Schroff2015} proposed a modification suited for retrieval tasks \emph{i.e.} equation \eqref{eq:triplet}, termed:
triplet loss. Triplet loss forces the data points from the same class to be closer to each other than a
data point form any other class. Notice that contrary to contrastive loss in \eqref{eq:contrastive}, triplet loss adds
context to the loss function by considering both a positive and negative pair distances from the same point.
\begin{equation}
  {l}_{triplet}(a,p,n) = [D_{ap} - D_{an} + \alpha]_+
  \label{eq:triplet}
  \end{equation}

Training complexity of triplet loss is $O(N^3)$ which is computationally prohibitive. High computational complexity of
triplet and contrastive losses have motivated a host of sampling approaches for an efficient optimization.

\subsection*{Dataset Sampling}

\begin{table*}[!t]
\centering
\begin{tabular}{|l| |c| |c| |c|}
	\hline
  \textbf{Sampling variant} & \textbf{Positive weight: $w_p$} & \textbf{Negative weight: $w_n$} & \textbf{Comments} \\
  \hline
  \hline
  Batch all (BA) & 1 & 1 & Uniformly weighted \\
  \hline
  Batch hard (BH) & $[~x_p ==  {\arg \underset{{x \in P(a)}}\max~~{D_{ax}}}~]$ & $[~x_n ==  {\arg \underset{{x \in N(a)}}\min~~{D_{ax}}}~]$  & Hardest sample \\
  \hline
  Batch sample (BS) & $[~x_p == \underset{x \in P(a)} {\textrm{multinomial}}~\{ D_{ax}\}~] $ & $[~x_n == \underset{x \in N(a)} {\textrm{multinomial}}~\{ -D_{ax}\}~]\footnotemark $ & Multinomial sampling \\
  \hline
  Batch weighted (BW) & $\dfrac{e^{D_{ap}}}{\underset{x \in P(a)} \sum e^{D_{ax}}}$ & $\dfrac{e^{-D_{an}}}{\underset{x \in N(a)} \sum e^{-D_{ax}}}$ & Adaptive weights \\
  \hline

\end{tabular}

\vspace{2mm}
\caption{Various ways of mining good samples in a batch, for better optimization of embedding loss.}
\label{tab:losses}
\end{table*}
\footnotetext{The conference version has a typo in BS: missing negative sign for negative weight \emph{c.f.} Table~\ref{tab:losses}}

As triplet and contrastive losses are computationally prohibitive for practical datasets, most proposed approaches resort to
sampling \emph{effective} data points for computing losses. This is important as computing loss over
trivial data points could only impair convergence of the algorithm. In the context of vehicles, it will be more informative
for a loss function to sample from different views (\emph{e.g.} side or front-view) for the same identity,
than considering samples of similar views repeatedly.

A popular sampling approach to find informative samples is \emph{hard data mining}, and is employed in many computer vision
applications \emph{e.g.} object detection. Hard data mining is a bootstrapping technique which is used in iterative
training of a model, wherein at every iteration the current model is applied on a validation set to mine hard data on which this model is performing poorly. Only these hard data are then presented to the optimizer which increases the ability of the model to learn effectively and
converge faster to an optimum.  On the flip side, if a model is only presented with hard data, which could comprise outliers, its
ability to discriminate outliers \emph{w.r.t.} normal data would suffer.

In order to deal with the outliers during hard data sampling, facenet \cite{Schroff2015} proposed \emph{semihard} sampling which mines moderate triplets that are neither too
hard nor too trivial for getting meaningful gradients during training. The generation of semihard samples is
performed offline and on CPU which severely impedes convergence. \cite{Hermans2017} proposed a very efficient and
effective approach to mine samples directly on GPU. The authors construct a data batch by randomly sampling $P$
identities from $\mathcal{X}$ and then randomly sampling $K$ images for each identity, thus resulting in a batch size of $PK$ images. In a
batch size of $PK$ images, the authors \cite{Hermans2017} proposed two sampling techniques, namely \textbf{batch hard} (BH)
(also in \cite{Mishchuk2017}) and
\textbf{batch all} (BA). Another sampling technique
\textbf{batch sample} (BS) is actively discussed in the implementation webpage of \cite{Hermans2017}, however to the best of our knowledge we could not find a formalized study and evaluation for this sampling technique.

\cite{Ristani2018} unifies different batch sampling techniques in \cite{Hermans2017} under one expression. Let $a$ be an anchor sample
and $N(a)$ and $P(a)$ represent a subset of negative and positive samples for the corresponding anchor $a$. The triplet
loss can then be written as:
\begin{equation}
  {l}_{triplet}(a) = [\alpha + \sum_{p \in P(a)}{w_p D_{ap}} - \sum_{n \in N(a)}{w_n D_{an} }]_+
  \label{eq:unified_loss}
\end{equation}

With respect to an anchor sample $a$: $w_p$ represents the
weight (importance) of positive sample $p$ and similarly $w_n$ signifies the
importance of the negative sample $n$.

  The total loss in an epoch is then obtained by:
  \begin{equation*}
  \mathcal{L(\theta; X)} = \sum_{all~batches} \sum_{a \in B} {l_{triplet}(a)}
  \end{equation*}

Table \ref{tab:losses} summarizes different ways of sampling positives and negatives. We formalize BS
method in this regime. BH is hard data
mining in the batch, using only the hardest positive and negative samples for every anchor. BA is a straightforward
sampling which gives uniform weights to all samples. Uniform weight distribution can ignore the
contribution of important tough samples as these samples are typically outnumbered by the trivial easy samples. In order to mitigate this issue with BA, \cite{Ristani2018} employs a weighting scheme \textbf{batch weighted} (BW), wherein a sample is weighted based on its distance from the corresponding anchor, thereby giving more importance to the informative harder samples than trivial samples.


BS uses the distribution of anchor-to-sample distances to mine
a positive and negative data for an anchor. This technique thereby avoids sampling outliers when compared with BH, and also hopes
to find out the most relevant sample as the sampling is done using distances-to-anchor distribution.

In the following sections, we evaluate the embedding losses, along with the sampling variants presented in Table~\ref{tab:losses}.

\section{Experiments}\label{sec:experiments}
For our evaluation purposes we use three popular  publicly available datasets: VeRi, VehicleID and PKU-VD. \\
\noindent \textbf{VeRi}: This dataset is proposed by \cite{Liu2016a} and is one of the main datasets used in vehicle
re-identification literature for comparative study. This dataset encompasses 40,000 bounding box annotations of 776 cars (identities) across 20 cameras in
traffic surveillance scenes. Each vehicle is captured in 2-18 cameras in various viewpoints and varying
illuminations. Notably the viewpoints are not restricted to only front/rear but also side views, thereby making it one
of the challenging datasets. The annotations include make and model of vehicles, color and inter-camera
relations and trajectory information.\\
\noindent \textbf{VehicleID}: This dataset \cite{Liu2016b} comprises 221,763 bounding boxes of 26,267 identities, captured across
various surveillance cameras in a city. Annotations include 250 vehicle models and this dataset has an order of magnitude more images than VeRi dataset. However the viewpoints only include front and rear views for
vehicles. \\
\noindent \textbf{PKU-VD}: \cite{Yan2017} proposed a large dataset for fine grained vehicle analysis
including re-identification and classification. To this date this is the largest dataset comprising about
\emph{two million} images and their fine grained labels including vehicle model and color. This dataset is split into two
sub-datasets, namely \textbf{VD1} and \textbf{VD2} based on cities from which they were captured. The images in VD1 are
captured from high resolution cameras, while images for VD2 are obtained from surveillance cameras. There are about 71k
and 36k identities in VD1 and VD2, respectively.

\subsection{Training and Hyperparameters}
For our experiments, we fix our backbone or meta-architecture to \emph{mobilenet} \cite{Howard2017} owing to its better efficiency
(parameters, speed) as compared to ResNet-variants \cite{He2016} and VGG \cite{Simonyan2014a}. The imagenet
\cite{JiaDeng2009} retrieval accuracy for these architectures are in similar ranges.

We use Adam optimizer \cite{Kingma2014} with default hyperparameters ( $\epsilon=10^{-3}$, $\beta_1 = 0.9$, $\beta_2 = 0.999$ ). Depending upon if the training is done from
scratch or fine-tuned using an imagenet \cite{JiaDeng2009} based trained model, we employ different learning rate
schedulers. When training from scratch, we use standard learning rate of $0.001$. We reduce this rate to
$0.0003$ when using an imagenet based pre-trained model. For online data augmentation a standard image-flip operation is used. We use Nvidia's Volta GPU for hardware and Tensorflow \cite{tensorflow2015-whitepaper} as the software platform.

We replace the margin $\alpha$  in triplet loss \eqref{eq:unified_loss}
by \emph{softplus} function: $ln(1 + exp(\cdot))$ which avoids the need of tuning this margin \cite{Hermans2017}. For contrastive loss we follow standard practice of hard margin of $1.0$. Using a softplus function produced poorer results for contrastive loss.

For the \emph{batch construction}, unless otherwise specified, we follow the default batch sizes as in
\cite{Hermans2017,Ristani2018}. A batch consists of $18$ (P) randomly chosen identities, and for each identity, $4$ (K) samples are chosen randomly, thereby selecting a total of $72$ (PK) images. Samples are chosen such that we iterate over all
train set during the course of an epoch. Following the standards in face-verification and
person re-identification \cite{Ristani2018}, \cite{Schroff2015} we set the embedding dimension to \emph{128
units}.

\subsection{Evaluation Metrics}
We use mean-average-precision (\emph{mAP}) and \emph{top-k} accuracy for evaluating and comparing our presented
approaches. In a typical re-identification evaluation setup, we have a query set and a gallery set. For each vehicle in
a query set the aim is to retrieve a similar identity from the test set (\emph{i.e.} gallery set). $AP(q)$ for a query
image $q$ is defined as:
\begin{equation*}
  AP(q) = \dfrac{\underset{k} {\sum} P(k) \times \delta_k}{N_{gt}(q)}
  \end{equation*}
where P(k) represents precision at rank $k$, $N_{gt}(q)$ is the total number of true retrievals for $q$. $\delta_k$ is 1
when the matching of query image $q$ to a test image is correct at rank $<= k$. $mAP$ is then computed as average over
all query images:
\begin{equation*}
mAP = \dfrac{\underset{q} {\sum} {AP(q)}} {Q}
\end{equation*}

where $Q$ is the total number of query images.

\section{Results and Discussions}\label{sec:results}
We present our results on the datasets mentioned in the previous section. Different datasets have different ways of
constructing test sets which
we elaborate in the respective sections. Each model presented below is trained separately on
the corresponding dataset using its standard train set.

\subsection{VeRi}\label{subsec:veri}
We follow the standard evaluation protocol by \cite{Wang2018}. The total number of query images is 1,678 while the
gallery set
comprises 11,579 images. For every query image, the gallery set contains images of same query-identity but taken from \emph{different}
cameras. This is an important evaluation exclusion as in many cases the same camera samples would contain visually similar samples for the same vehicle.

\begin{table}[!h]
\centering
\begin{tabular}{|l| |c| |c| |c| |c|}
  \hline
  \textbf{Sampling} & \textbf{mAP} & \textbf{top-1} & \textbf{top-2} & \textbf{top-5} \\
  \hline
  \multicolumn{5}{|c|}{\textbf{Triplet, Not-Normalized}} \\
  \hline
  BH & 65.10 & 87.25 & 91.54 & 94.76 \\
  \hline
  BA & 66.91 & 90.11 & \textbf{93.38} & 96.01 \\
  \hline
  BS & \textbf{67.55} & \textbf{90.23} & 92.91 & 96.42 \\
  \hline
  BW & 67.02 & 89.99 & 93.15 & \textbf{96.54} \\
  \hline

  \multicolumn{5}{|c|}{\textbf{Triplet, Normalized}} \\
  \hline
  BH & 53.72 & 72.65 & 80.27 & 86.83 \\
  \hline
  BA & 27.60 & 42.91 & 53.16 & 67.76 \\
  \hline
  BS & 33.79 & 48.75 & 58.64 & 73.54 \\
  \hline
  BW & 44.29 & 60.91 & 69.85 & 80.63 \\
  \hline
  \multicolumn{5}{|c|}{\textbf{Contrastive, Normalized}} \\
  \hline
  BH & 59.21 & 80.51 & 85.52 & 90.64 \\
  \hline
  BS & 52.09 & 71.51 & 78.84 & 86.95 \\
  \hline
  \multicolumn{5}{|c|}{\textbf{Contrastive, Not-Normalized}} \\
  \hline
  BH & 56.84 & 75.33 & 82.30 & 90.29 \\
  \hline
  BS & 48.85 & 65.49 & 74.55 & 85.76 \\
  \hline
  
\end{tabular}
\vspace{1mm}
\caption{VeRi accuracy results (\%) using triplet and contrastive loss for different batch sampling variants outlined in Table \ref{tab:losses}.}
\label{tab:veri}
\end{table}

Table \ref{tab:veri} summarizes our results for various sampling configurations, and we can draw following inferences:
\begin{itemize}
\item Adding a normalized layer performs poorly for the triplet loss. This is also reported by \cite{Hermans2017} wherein using a normalized layer
  could result in collapsed embeddings.
\item Siamese (contrastive) loss under performs relative to triplet loss. We attribute this to the additional context
  provided by using both positive and negative samples in the same term for the triplet loss \cite{Manmatha2017}.
\item For the best performing set, \emph{i.e.} triplet loss with no-normalization layer: all four sampling variants reach about similar accuracy ranges, with BS outperforming others in a close range.
\item Figure \ref{fig:veri_results} shows some visual results with embeddings learned from batch-sampling triplet
  loss. Good top-k retrievals indicate stability of our embeddings across different views and cameras. Notice that
  query and gallery images are constrained to be from different cameras following the standard evaluation protocol.
\end{itemize}

\noindent \textbf{Comparison to the state-of-the-art approaches}: Table \ref{tab:veri_sota} outlines comparisons with the state-of-the-art approaches. Notice that our approach outperforms all the other results for the \emph{mAP}
metric. GSTE \cite{Bai2018} achieves better top-k accuracy but in terms of mAP our approach performs better indicating robustness at all
ranks. Furthermore GSTE \cite{Bai2018} has an embedding dimension of 8x more (\emph{i.e.} 1024) than ours, and GSTE
includes a complicated training process which requires tuning an additional intra-class clustering parameter.

\begin{table}[!h]
\centering
\begin{tabular}{|l| |c| |c| |c|}
  \hline
  \textbf{Method} & \textbf{mAP} & \textbf{top-1} & \textbf{top-5} \\
  \hline
  BS (Ours) & \textbf{67.55} & 90.23 & 96.42 \\
  \hline
  GSTE \cite{Bai2018} & 59.47 & \textbf{96.24}  & \textbf{98.97} \\
  \hline
  VAMI \cite{Zhou2018a} & 50.13 & 77.03  & 90.82 \\
  \hline
  VAMI+ST * \cite{Zhou2018a} & 61.32 & 85.92  & 91.84 \\
  \hline
  OIFE \cite{Wang2017} & 48.00 & 89.43  & - \\
  \hline
  OIFE+ST *\cite{Wang2017} & 51.42 & 92.35 & - \\
  \hline
  PROVID * \cite{Wang2018} & 27.77 & 61.44 & 78.78 \\
  \hline
  Path-LSTM * \cite{Shen2017} & 58.27 & 83.49 & 90.04 \\
  \hline
\end{tabular}
\vspace{1mm}
\caption{\textbf{Comparison} of various proposed approaches on VeRi dataset. (*) indicates the usage of spatio-temporal
  information.}
\label{tab:veri_sota}
\end{table}

The VeRi dataset includes spatio-temporal (ST) information and \cite{Zhou2018a,Zhou2018,Wang2017,Wang2018}
utilize ST information in either embedding or in retrieval stages. \emph{Noticeably} without using any ST information,  we outperform these
approaches using ST. Contrary to us, OIFE \cite{Wang2017} requires extra annotations of
keypoints during training for their orientation invariant embedding learning. Training procedure for VAMI
\cite{Zhou2018a} include \emph{generative adversarial network} (GAN) and multi-view attention learning. Path-LSTM \cite{Shen2017} employ generation of several
path-proposals for their spatio-temporal regularization and requires an additional LSTM to rank these proposals.
It is worth noting that our training procedure is more straightforward than most of the approaches presented in Table~\ref{tab:veri_sota}, with an efficient embedding dimension of
128. Table~\ref{tab:comp_hypers} outlines some important differences \emph{w.r.t.} competitive approaches.

\begin{table}[!h]
  \centering
  \begin{tabular}{|l |c |c |c|}
    \hline
    \textbf{Method} & \textbf{ED} & \textbf{Multi-View} & \textbf{Annotations} \\
    \hline
    Ours & 128 & No & ID \\
    \hline
    GSTE \cite{Bai2018} & 1024 & No & ID \\
    \hline
    VAMI \cite{Zhou2018a} & 2048 & Yes & ID + Attribute \\
    \hline
    OIFE \cite{Wang2017} & 256 & No & ID + Keypoints \\
    \hline
    MGR \cite{Yan2017} & 1024 & No & ID + Attribute \\
    \hline
    ATT \cite{Yan2017} & 1024 & No & ID + Attribute \\
    \hline
    C2F \cite{Guo2018} & 1024 & No & ID + Attribute \\
    \hline
    CLVR \cite{Kanaci2017} & 1024 & No & Attribute \\
    \hline
  \end{tabular}
  \vspace{1mm}
  \caption{Summary of some important hyperparameters and labeling used during training. \textbf{ED} indicates embedding
    dimension. OIFE merges four datasets to form one large training set. Notice that our ED is the least among other approaches.}
  \label{tab:comp_hypers}
\end{table}

Referring to the best results in Table~\ref{tab:veri}, in the subsequent sections we consider only triplet
loss without embedding-normalization.

\subsection{PKU-VD}
PKU-VD is a large dataset combining two sub-datasets, VD1 and VD2. Both of these comprise about 400k training
images. The test set of each of the sub-dataset is split
into \emph{three} reference sets: small, medium and large. Table~\ref{tab:pku_testset} shows the number of test images in each
sub-dataset. For evaluation, we use the same dataset files for each reference set as provided by the authors \cite{Yan2017} of this dataset.
\begin{table}[!h]
  \centering
  \begin{tabular}{|l| |c| c| c| c|}
    \hline
    \textbf{Dataset} & \textbf{Small} & \textbf{Medium} & \textbf{Large} \\
    \hline
    VD1 & 106,887 & 604,432 & 1,097,649 \\
    \hline
    VD2 & 105,550 & 457,910 & 807,260 \\
    \hline
  \end{tabular}
  \vspace{1mm}
  \caption{Number of images in each reference test-set.}
  \label{tab:pku_testset}
\end{table}

Compared to VeRi and VehicleID datasets, PKU-VD dataset has an order of magnitude more images, hence a network can be trained from scratch
on this dataset. Furthermore with more intra-class samples, one can increase the batch size of triplets. Tables \ref{tab:VD_scratch}, \ref{tab:VD_scratch_all_results} and \ref{tab:VD_pretrained_default} show results for various configurations. 
For the BW sampling in Table~\ref{tab:VD_scratch}, the numerics following illustrate the \emph{P} and \emph{K} values, described previously, which create the batch. Table~\ref{tab:VD_scratch_all_results} adds results for the other three sampling variants when training from scratch. Table~\ref{tab:VD_pretrained_default} shows results for the default batch size (18x4).

Using more triplets in the batch improves the
accuracy, which is intuitively satisfying. Noticeably using the hardest sample (BH) does not kick-off the training (\emph{c.f.} Table~\ref{tab:VD_scratch_all_results}). This is expected and also noted in \cite{Schroff2015}, as with BH due to random initialization, the network never learns any understanding to separate hard data from easy samples. One way to deal with this is to start training with a few identities in a multi-class setting in-order to pre-train the network and then proceed with the standard BH procedure. Alternatively one could start from an imagenet trained network (\emph{c.f.} Table~\ref{tab:VD_pretrained_default}). The other sampling variants are more robust and hence converges to a better solution than the default $PK$ batch-sized training.

BW sampling with bach size of 18x16 outperforms the precious state-of-the-art by \cite{Yan2017}. Multi-grain ranking (MGR) uses permutation probability based ranking method
and include vehicle attributes during training process. Noticeably our training procedure is straightforward without using
vehicle attributes. Furthermore MGR uses an embedding dimension of 1024 as opposed to 128 for our embedding, thus calling for higher computation cost during inference in \cite{Yan2017}.

\begin{table}[h]
\centering
\begin{tabular}{|l| |c| c| c| c|}
  \hline
  \textbf{Method} & \textbf{Small} & \textbf{Medium} & \textbf{Large} \\
  \hline
  \multicolumn{4}{|c|}{\textbf{VD1}} \\
  \hline
  BW (18x16) & \textbf{87.48} & \textbf{67.28} & \textbf{58.77}  \\
  \hline
  MGR \cite{Yan2017} & 79.10 & 58.30 & 51.10 \\
  \hline
  \multicolumn{4}{|c|}{\textbf{VD2}} \\
  \hline
  BW (18x16) & \textbf{84.55} & \textbf{69.87} & \textbf{63.64} \\
  \hline
  MGR \cite{Yan2017} & 74.70 & 60.60 & 55.30 \\
  \hline
\end{tabular}
 \vspace{1mm}
\caption{\textbf{mAP} (\%) for retrievals on various reference sets. Training is performed from scratch without using
  pretrained weights.}
\label{tab:VD_scratch}
\end{table}

\begin{table}[!h]
\centering
\begin{tabular}{|l| |c| c| c| c|}
  \hline
  \textbf{Dataset, Sampling} & \textbf{Small} & \textbf{Medium} & \textbf{Large} \\
  \hline
  \multicolumn{4}{|c|}{\textbf{No pretrained weights}} \\
  \hline
  VD1, BA  & 85.02 & 62.84 & 54.68  \\
  \hline
  VD1, BH  & 0.00 & 0.00 & 0.00  \\
  \hline
  VD1, BS & 87.24 & 66.62 & 58.26 \\
  \hline
  VD2, BA & 83.39 & 68.58 & 62.34 \\
  \hline
  VD2, BH  & 0.00 & 0.00 & 0.00  \\
  \hline
  VD2, BS & 83.30 & 68.45 & 62.36 \\
  \hline
\end{tabular}
\vspace{1mm}
\caption{\textbf{mAP} (\%) for retrievals on various reference sets of different sizes. Training is performed without
  pretrained weights with batch size of 18x16.}
\label{tab:VD_scratch_all_results}
\end{table}

\begin{table}[!h]
\centering
\begin{tabular}{|l| |c| c| c| c|}
  \hline
  \textbf{Dataset, Sampling} & \textbf{Small} & \textbf{Medium} & \textbf{Large} \\
  \hline
  \multicolumn{4}{|c|}{\textbf{With pretrained weights}} \\
  \hline
  VD1, BW & \textbf{82.66} & 60.15 & 52.10 \\
  \hline
  VD1, BS  & 81.36 & 58.91 & 50.68  \\
  \hline
  VD1, BA  & 79.46 & 56.79 & 49.26  \\
  \hline
  VD1, BH  & 82.04 & \textbf{60.40} & \textbf{52.17}  \\
  \hline
  VD2, BW & \textbf{80.93} & \textbf{65.44} & \textbf{58.94} \\
  \hline
  VD2, BS & 75.52 & 58.35 & 51.71 \\
  \hline
  VD2, BA & 70.07 & 50.56 & 43.46 \\
  \hline
  VD2, BH  & 78.95 & 62.32 & 55.86  \\
  \hline
\end{tabular}
\vspace{1mm}
\caption{\textbf{mAP} (\%) for retrievals on various reference sets of different sizes. Training is performed using
  imagenet pretrained weights with default batch size of 18x4.}
\label{tab:VD_pretrained_default}
\end{table}

\subsection{VehicleID}
VehicleID \cite{Liu2016b} is a larger dataset than VeRi containing front and rear views for the vehicles. We follow the standard evaluation protocol of \cite{Liu2016b} and
provide results on four reference \emph{query} sets. Reference sets: small, medium. large and X-large contain 800, 1600, 2400 and
13164 identities, respectively. For each reference set, an exemplar for an identity is randomly chosen, and a \emph{gallery}
set is constructed. This process is repeated ten times to obtain
\emph{averaged} evaluation metrics. For training we use
mobilenet network, pretrained using imagenet dataset, without normalization-layer for embedding. Similarly to the PKU-VD dataset training we set the batch size
($PK$) to 18x16 images. For the sake of completeness we provide the results with default PK batch size of 18x4.

\begin{table}[!h]
  \centering
  \begin{tabular}{|l |c |c |c |c|}
    \hline
    \textbf{Method} & \textbf{Small} & \textbf{Medium} & \textbf{Large} & \textbf{X-Large} \\
    \hline
    BA & 84.65 & 79.85 & 75.95 & 59.74 \\
    \hline
    BS & \textbf{86.19} & \textbf{81.69} & \textbf{78.16} & \textbf{62.41} \\
    \hline
    BW & 85.92 & 81.41 & 78.13 & 62.12 \\
    \hline
    BH & 85.59 & 80.76 & 76.87 & 60.33 \\
    \hline
    C2F \cite{Guo2018} & 63.50 & 60.00 & 53.00 & - \\
    \hline
    GSTE \cite{Bai2018} & 75.40 & 74.30 & 72.40 & - \\
    \hline
    ATT \cite{Yan2017} & 62.80 & 62.30 & 58.60 & - \\
    \hline
    CCL \cite{Liu2016b} & 54.60 & 48.10 & 45.50 & - \\
    \hline
  \end{tabular}
  \vspace{1mm}
  \caption{Accuracy results on VehicleID using \textbf{mAP} metric (\%). Batch size for our experiments is set to
    $18$x$16$ samples.}
  \label{tab:vid_map}
\end{table}

\begin{table}[!h]
  \centering
  \begin{tabular}{|l |c |c |c |c|}
    \hline
    \textbf{Method} & \textbf{Small} & \textbf{Medium} & \textbf{Large} & \textbf{X-Large} \\
    \hline
    BA & 81.90 & 76.57 & 72.60 & 54.95 \\
    \hline
    BS & 84.17 & 79.05 & 75.52 & 59.10 \\
    \hline
    BW & \textbf{84.90} & \textbf{80.80} & \textbf{77.20} & \textbf{60.92} \\
    \hline
    BH & 83.34 & 78.72 & 75.02 & 57.97 \\
    \hline
  \end{tabular}
  \vspace{1mm}
  \caption{Accuracy results on VehicleID using \textbf{mAP} metric (\%). This is with default PK batch size of (18x4).}
  \label{tab:vid_map_default}
\end{table}

Tables~\ref{tab:vid_map},~\ref{tab:vid_map_default}~and~\ref{tab:vid_topk} show comparative results for mAP and top-k metrics,
respectively. Similarly to the PKU-VD results, using a larger batch size increases the retrieval rankings, however the margin of improvement is smaller. This could be due to limited variability in this dataset in terms of viewpoints and number of vechicle-models, owing to which increasing the batch size does not necessarily increase informative statistics.

\begin{table}[!h]
  \centering
  \begin{tabular}{|l |c |c |c |c|}
    \hline
    \textbf{Method} & \textbf{Small} & \textbf{Medium} & \textbf{Large} & \textbf{X-Large} \\
    \hline
    \multicolumn{5}{|c|}{\textbf{Top-1}} \\
    \hline
    BA & 76.69 & 71.20 & 66.71 & 50.22 \\
    \hline
    BS & \textbf{78.80} & 73.41 & 69.33 & \textbf{53.07} \\
    \hline
    BW & 78.49 & 73.10 & 69.41 & 52.82 \\
    \hline
    BH & 77.90 & 72.14 & 67.56 & 50.67 \\
    \hline
    OIFE \cite{Wang2017} & - & - & 67.00 & - \\
    \hline
    OIFE+ \cite{Wang2017} & - & - & 68.00 & - \\
    \hline
    VAMI \cite{Zhou2018a} & 63.12 & 52.87 & 47.34 & - \\
    \hline
    CCL \cite{Liu2016b} & 49.00 & 42.80 & 38.20 &  -\\
    \hline
    C2F \cite{Guo2018} & 61.10 & 56.20 & 51.40 & - \\
    \hline
    GSTE \cite{Bai2018} & 75.90 & \textbf{74.80} & \textbf{74.00} & - \\
    \hline
    CLVR \cite{Kanaci2017} & 62.00 & 56.10 & 50.60 & - \\
    \hline
    \multicolumn{5}{|c|}{\textbf{Top-5}} \\
    \hline
    BA & 95.26 & 91.17 & 87.75 & 70.48 \\
    \hline
    BS & \textbf{96.17} & \textbf{92.57} & \textbf{89.45} & \textbf{73.06} \\
    \hline
    BW & 95.83 & 92.48 & 89.36 & 72.72 \\
    \hline
    BH & 95.74 & 92.03 & 88.81 & 71.23 \\
    \hline
    OIFE \cite{Wang2017} & - & - & 82.90 & - \\
    \hline
    VAMI \cite{Zhou2018a} & 83.25 & 75.12 & 70.29 & - \\
    \hline
    CCL \cite{Liu2016b} & 73.50 & 66.80 & 61.60 & -\\
    \hline
    C2F \cite{Guo2018} & 81.70 & 76.20 & 72.20 & - \\
    \hline
    GSTE \cite{Bai2018} & 84.20 & 83.60 & 82.70 & - \\
    \hline
    CLVR \cite{Kanaci2017} & 76.00 & 71.80 & 68.00 & - \\
    \hline
  \end{tabular}
  \vspace{1mm}
  \caption{Results on VehicleID dataset using \textbf{top-k} metric (\%). Batch size for our experiments is set to
    $18$x$16$ samples.}
  \label{tab:vid_topk}
\end{table}

BS and BW outperform other sampling variants, including all
state-of-the-art approaches in the mAP metric. 
Table~\ref{tab:comp_hypers} and section~\ref{subsec:veri} summarizes important differences of
state-of-the-art approaches \emph{w.r.t.} our approach. GSTE \cite{Bai2018} achieves better
performance in terms of top-1 accuracy, but their accuracy drops for top-5. Lower mAP and top-5 indicates GSTE's sub-par retrieval
performances for ranks $k>1$.
OIFE+ \cite{Wang2017} achieves close accuracy in top-5 to ours. As opposed to our approach, OIFE+ requires
keypoint annotations and a separate metric learning module from \cite{Zhang2016}. Furthermore OIFE combines VeRi,
VehicleID, CompCars \cite{Luo2015} and
Cars21k \cite{Sochor2016} into one large train set.

Contrary to our method, other approaches
\cite{Guo2018,Liu2016b,Yan2017}, all utilize model annotations (in addition to identity annotations) from the training set for
re-identification.

\begin{figure*}[h]
  \centering
  \includegraphics[width=0.999\textwidth]{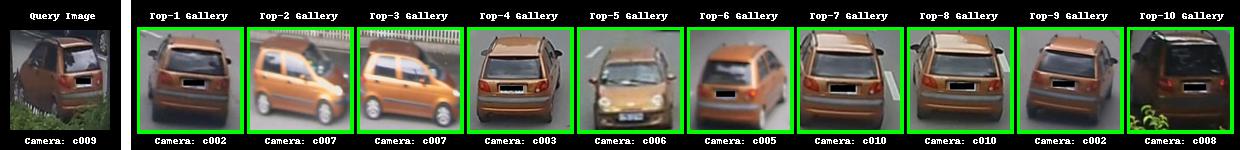}
  \\
  \includegraphics[width=0.999\textwidth]{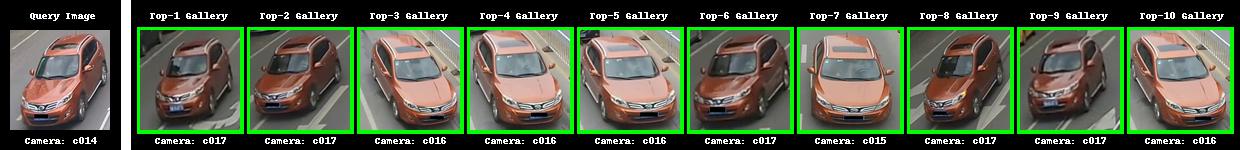}
  \\
  \includegraphics[width=0.999\textwidth]{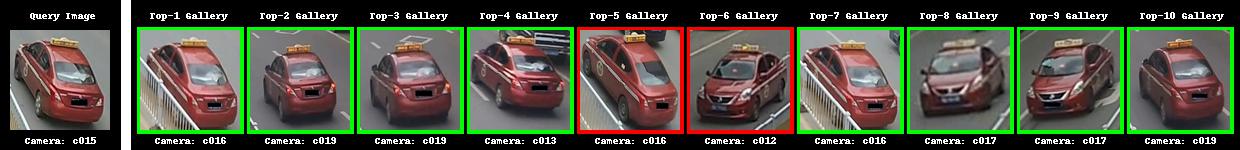}
  \\
  \includegraphics[width=0.999\textwidth]{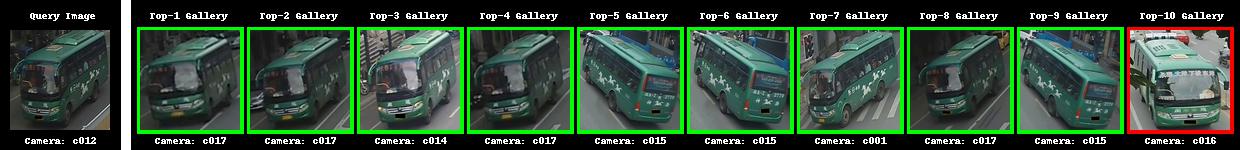}
  \\
  \includegraphics[width=0.999\textwidth]{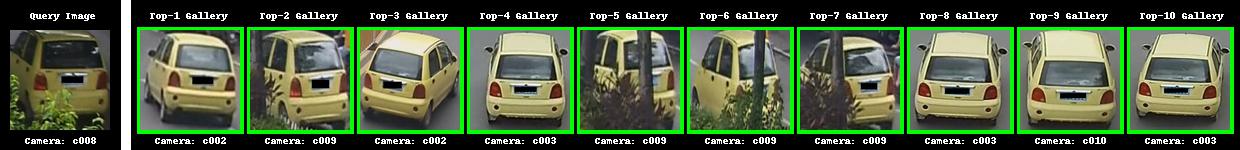}
  \\
  \includegraphics[width=0.999\textwidth]{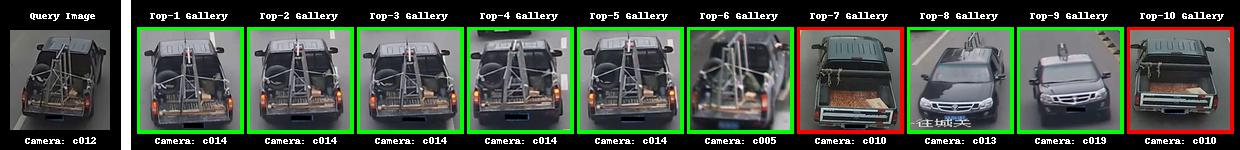}
  \\
  \includegraphics[width=0.999\textwidth]{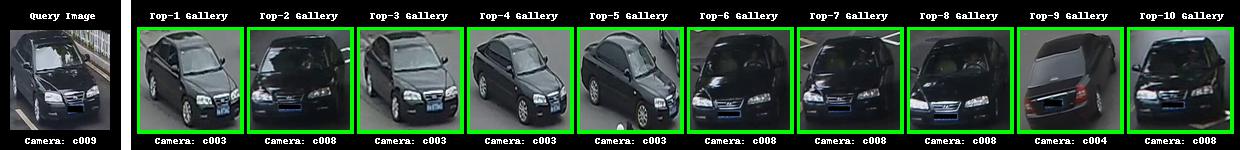}
  \\
  \includegraphics[width=0.999\textwidth]{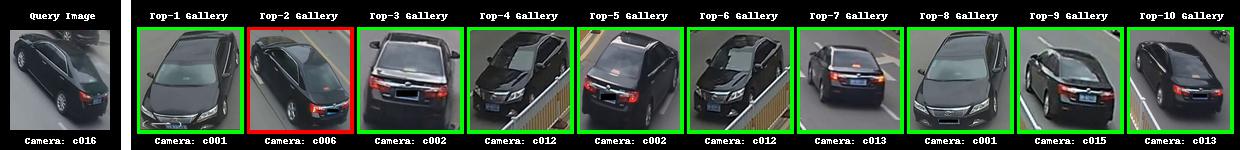}
  \\
  \includegraphics[width=0.999\textwidth]{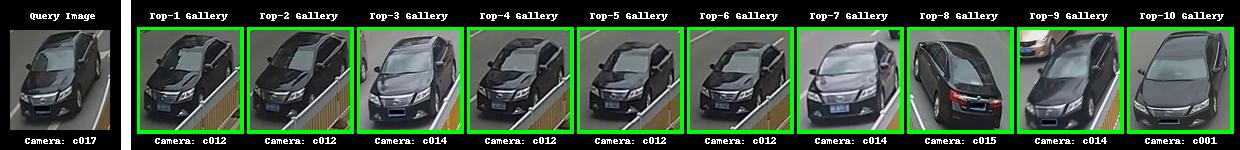}
  \\
  \includegraphics[width=0.999\textwidth]{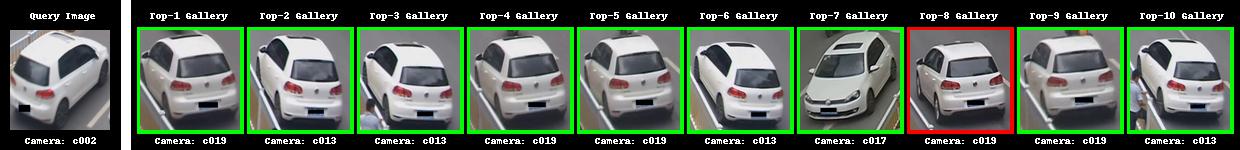}
  \caption{Qualitative results on VeRi dataset using \emph{BS} based triplet embedding. Each row indicates query image and top-10 retrievals for this query
    image. {\color{red}{Red}} border indicates incorrect retrieval and {\color{green}{Green}} indicates correct
    retrievals. These demonstrate good embedding quality as the top retrievals include different views and cameras. }
  \label{fig:veri_results}
\end{figure*}

\subsection{Cityflow dataset}
Results on CityFLow dataset \cite{Tang_2019_CVPR} proposed at CVPR 2019 can be found in \cite{Tang_2019_CVPR}.

\subsection{VRIC \protect \footnote{Experiment on this dataset is performed post IJCNN conference.}}
\cite{2018gcpr-Kanaci} proposed this dataset comprising large variations in scale, motion, illumination, occlusion and viewpoint. This set contains 60,430 images of 5,622 vehicle identities captured by 60 different cameras in both day-time and night-time. Following the standard in \cite{2018gcpr-Kanaci}, the query and probe splits are set to 2811 identities. Evaluation results from various sampling techniques are presented in table~\ref{tab:vric}.
\begin{table}[!h]
\centering
\begin{tabular}{|l| |c| |c| |c|}
  \hline
  \textbf{Method} & \textbf{mAP} & \textbf{top-1} & \textbf{top-5} \\
  \hline
  BA & {75.11} & 64.18 & 89.4 \\
  \hline
  BH & 77.99 & {67.77}  & {91.32} \\
  \hline
  BS & 76.78 & 66.83  & 90.64 \\
  \hline
  BW & 78.55 & 69.09  & 90.54 \\
  \hline
  \cite{2018gcpr-Kanaci} & - & 46.61  & 65.58 \\
  \hline
\end{tabular}
\vspace{1mm}
\caption{\textbf{Comparison} of various approaches on Vric dataset. We use the default standard network training sceheme - pretrained weiughts from imagenet, mobilenet-v1 as the feature extractor and batch size PK is set to 18x4 images (embedding dimension at 128). }
\label{tab:vric}
\end{table}
As shows in table~\ref{tab:vric}, our standard experimental settings outperform the baseline multi-scale matching approach by the dataset authors \cite{2018gcpr-Kanaci}.

\subsection{Veri-Wild\protect\footnotemark[3]}
Veri-Wild \cite{lou2019large} is the largest dataset as of CVPR 2019. The dataset is captured from a large CCTV surveillance system consisting of 174 cameras across one month (30× 24h) under unconstrained scenarios. This dataset comprises 416,314 vehicle images of 40,671 identities.
Evaluation on this dataset is split across three subsets: small, medium and large; comprising 3000, 5000 and 10,000 identities respectively (in probe and gallery sets). 

Table~\ref{tab:vwild} demonstrates our results on this dataset. 
\begin{table}[!h]
  \centering
  \begin{tabular}{|l |c |c |c|}
    \hline
    \textbf{Method} & \textbf{Small} & \textbf{Medium} & \textbf{Large} \\
    \hline
    \multicolumn{4}{|c|}{\textbf{Top-1}} \\
    \hline
    BA & 82.83 & 78.06 & 69.72  \\
    \hline
    BH & 83.30 & 76.90 & 69.10  \\
    \hline
    BS & 82.90 & 77.68 & 69.59  \\
    \hline
    BW & 84.17 & 78.22 & 69.99  \\
    \hline
    \cite{lou2019large} & 64.03 & 57.82 & 49.43 \\
    \hline
    \hline
    \multicolumn{4}{|c|}{\textbf{Top-5}} \\
    \hline
    BA & 95.27 & 93.02 & 88.32 \\
    \hline
    BH & 95.20 & 92.66 & 87.74  \\
    \hline
    BS & 95.00 & 92.90 & 87.89  \\
    \hline
    BW & 95.30 & 93.06 & 88.45  \\
    \hline
    \cite{lou2019large} & 82.80 & 78.34 & 70.48 \\
    \hline
    \hline
    \multicolumn{4}{|c|}{\textbf{mAP}} \\
    \hline
    BA & 68.21 & 60.69 & 49.28 \\
    \hline
    BH & 69.37 & 61.47 & 50.27  \\
    \hline
    BS & 68.79 & 61.11 & 49.79  \\
    \hline
    BW & 70.54 & 62.83 & 51.63  \\
    \hline
    \cite{lou2019large} & 35.11 & 29.80 & 28.78 \\
    \hline
  \end{tabular}
  \vspace{1mm}
  \caption{Results on Veri-Wild dataset using \textbf{top-k} metric (\%) and \textbf{mAP}. Batch size for our experiments is set to
    $18$x$4$ samples. Default experimental settings were used (embedding dimension at 128).}
  \label{tab:vwild}
\end{table}

\section{Conclusion and Future Work}
In this paper we propose a strong baseline for vehicle re-identification using the best practices in learning deep triplet
embedding \cite{Hermans2017}. The core ideas behind this set of best practices lie in constructing a batch to facilitate extracting meaningful statistics in order to guide training and convergence.
We also introduced a formal exposition and evaluation of a triplet sampling variant, \emph{batch sample} to the re-identification literature.

We compared our baselines with the state-of-the-art approaches on three datasets and outperform almost all of
them in a wide range of evaluation criteria. 
The sampling variants: \emph{batch sample} and \emph{batch weighted} proved generally more effective and robust than \emph{batch hard} and \emph{batch all}.

We hinged our research on the belief that despite the intra-class variations, the identity of a vehicle is less fine
grained than other object re-identification task, \emph{e.g.} person re-identification.
Our results demonstrate this by using the recent advances in
embedding learning, we can push the frontiers of vehicle re-identification much further without using any spatio-temporal
information.
On the other hand, two vehicles of exactly the same color and
model (with subtle or no discerning marks, \emph{e.g.} last row in Figure~\ref{fig:veri_results}) would be very difficult to distinguish without any
spatio-temporal information. Incorporating spatio-temporal information along with other attributes in
an effective manner is
an important contribution as future work.

\clearpage

{\small
\bibliographystyle{ieee}
\bibliography{library}
}

\end{document}